\documentclass{article}

\usepackage[dblblindworkshop, final]{coml_2025}

\workshoptitle{Constrained Optimization for Machine Learning (COML)}

\usepackage[utf8]{inputenc} %
\usepackage[T1]{fontenc}    %
\usepackage{hyperref}       %
\usepackage{url}            %
\usepackage{booktabs}       %
\usepackage{amsfonts}       %
\usepackage{nicefrac}       %
\usepackage{microtype}      %
\usepackage[svgnames]{xcolor}         %

\usepackage{amsmath,amssymb,amsthm}
\usepackage{subcaption}

\usepackage{algorithm,algpseudocode}

\usepackage{tikz,pgfplots}
\usepackage{tkz-kiviat} 
\usetikzlibrary{arrows,positioning,fit}
\usepackage{tikzscale}

\def\addlegendimage{\csname pgfplots@addlegendimage\endcsname}

\usepackage{graphicx,mathtools}

\colorlet{MyBlue}{DodgerBlue!75!Black}
\colorlet{MyGreen}{DarkGreen!85!Black}
\colorlet{MyGray}{White!75!Black}

\definecolor{c1}{RGB}{238,102,119}
\definecolor{c2}{RGB}{68, 119, 170}
\definecolor{c3}{RGB}{102, 204, 238}
\definecolor{c4}{RGB}{34, 136, 51}
\definecolor{c5}{RGB}{204, 187, 68}
\definecolor{c6}{RGB}{170, 51, 119}
\definecolor{c7}{RGB}{187, 187, 187}

\definecolor{vibrant1}{HTML}{EE7733} %
\definecolor{vibrant2}{HTML}{0077BB} %
\definecolor{vibrant3}{HTML}{33BBEE} %
\definecolor{vibrant4}{HTML}{EE3377} %
\definecolor{vibrant5}{HTML}{CC3311} %
\definecolor{vibrant6}{HTML}{009988} %
\definecolor{vibrant7}{HTML}{BBBBBB} %

\usepackage{diagbox} %

\usepackage{siunitx} %

\usepackage{numprint}
\nprounddigits{3}

\hypersetup{
final,
colorlinks=true,
linktocpage=true,
pdfstartview=FitH,
breaklinks=true,
pdfpagemode=UseNone,
pageanchor=true,
pdfpagemode=UseOutlines,
plainpages=false,
bookmarksnumbered,
bookmarksopen=false,
bookmarksopenlevel=1,
hypertexnames=true,
pdfhighlight=/O,
urlcolor=Maroon,linkcolor=MyBlue!60!black,citecolor=DarkGreen!70!black,	%
pdftitle={},
pdfauthor={},
pdfsubject={},
pdfkeywords={},
pdfcreator={pdfLaTeX},
pdfproducer={LaTeX with hyperref}
}

\usepackage{cleveref}

\usepackage{booktabs}		%
\usepackage[inline,shortlabels]{enumitem}		%

\usepackage{pifont}%
\newcommand{\cmark}{\ding{51}}%

\usepackage{color-edits}
\addauthor{gb}{blue}
\addauthor{ja}{orange}
\addauthor{ak}{pink}

\newcommand{\cP}{\mathcal P}

\newcommand{\cX}{\mathcal X}

\newcommand{\bbE}{\mathbb E}

\newcommand{\bbR}{\mathbb R}

\newcommand{\R}{\bbR}

\newcommand{\C}{\mathcal{C}}

\DeclareMathOperator\proj{proj}

\newcommand{\iid}{\mathop{\sim}\limits^{iid}}

\def\rot{\rotatebox} %

\newcommand{\newmacro}[2]{\newcommand{#1}{#2}}		%

\newmacro{\step}{\gamma}		%
\newmacro{\learn}{\eta}		%

\newmacro{\ite}{k}
\newmacro{\initite}{0}
\newmacro{\afterinitite}{1}
\newmacro{\sumite}{n}
\newmacro{\Afterite}{K}

\def\SSW{SSw}

\title{humancompatible.train: Implementing Optimization Algorithms for Stochastically-Constrained Stochastic Optimization Problems}

\author{%
  Andrii Kliachkin\thanks{Equal contribution, more junior authors listed first.},~~ Jana Lep\v{s}ov\'a${}^{*}$,~~  Gilles Bareilles${}^{*}$, ~~ Jakub Marecek \\
  Artificial Intelligence Center, \\
  Czech Technical University in Prague \\
  \texttt{firstname.lastname@fel.cvut.cz}
}

\begin{document}

\maketitle

\begin{abstract}
There has been a considerable interest in constrained training of deep neural networks (DNNs) recently for applications such as fairness and safety. Several toolkits have been proposed for this task, yet there is still no industry standard. We present humancompatible.train (https://github.com/humancompatible/train), an easily-extendable PyTorch-based Python package for training DNNs with stochastic constraints. We implement multiple previously unimplemented algorithms for stochastically constrained stochastic optimization. We demonstrate the toolkit use by comparing two algorithms on a deep learning task with fairness constraints.
\end{abstract}

\section{Introduction}

There has been a considerable interest in constrained training of deep neural networks (DNNs), recently \cite{ramirez2025positionadoptconstraintspenalties}. 
Therein, one considers an empirical risk minimization (ERM) problem with constraints involving expectations:
\begin{equation}\label{eq:pb}
    \min_{x\in\R^n} \mathbb{E}[f(x,\xi)] \quad \text{s.t.} \quad \mathbb{E}[c(x,\zeta)] \leq 0,
\end{equation}
where $\xi$ and $\zeta$ are random variables, and $f(\cdot, \xi):\bbR^n \to \bbR$ and $c(\cdot, \zeta):\bbR^n\to\bbR^m$ are continuous nonsmooth nonconvex mappings.
This can be seen as a special case of stochastically constrained stochastic optimization problem, whose applications in operations research and statistics go well beyond DNNs.

In this work, we present \texttt{humancompatible.train}: a Python toolkit for training of neural networks with constraints as in \eqref{eq:pb}, and demonstrate it on a fairness use-case. 
This draws on numerous recent proposals  \cite{FaNaMaKo2024,BerCur2021,Curtis2024,Ozto2023,BerCur2023,NaAnKo2022,NaAnKo2023,Bolla2023,CurRob2024,ShiWaWa2022,FacKun2023,JakubReview,HuaLin2023}  to solve convex and non-convex empirical-risk minimization problems subject to constraints bounding the absolute value of empirical risk.
See Table \ref{table:algs_asm} for an overview. 
Numerous other algorithms of this kind could be construed, based on a number of design choices, including: (a) sampling techniques for the ERM objective and the constraints, either the same or different; (b)
 use of first-order or higher-order derivatives;
(c) use of globalization strategies such as filters or line search;
(d) use of ``true'' globalization strategies including random initial points and random restarts in order to reach global minimizers.

Nevertheless, there is no established toolkit, yet, which would implement all of these  algorithms and which would allow for their fair comparison.
This is not to say that there are no toolkits developed in this direction. 
Indeed, the package CHOP \cite{chop} implements Proximal Gradient Descent \cite{madry2018towards}, Frank-Wolfe \cite{locatello2019stochastic}, and the Stochastic Three-Composite Minimization \cite{yurtsever2016stochastic,pmlr-v139-yurtsever21a}, but is not actively maintained anymore.
These algorithms require the projection, which is non-trivial in training general DNNs, to be efficiently implementable.
There is the recently proposed Cooper \cite{gallegoPosada2025cooper}, that focuses on translating constrained problem to the unconstrained problem of minimizing the augmented Lagrangian. 
In contrast with unconstrained minimization of stochastic nonconvex objectives, where convergence guarantees exists \cite{davis2020stochastic}, no convergence guarantees are available for the Augmented Lagrangian method in the stochastic nonconvex objective and constraints setting; see SSL-ALM \cite{JakubReview}.
There is also GeoTorch \cite{lezcano2019trivializations}, 
which focuses on particular smooth manifolds encountered, e.g., in quantum information theory. 
While our toolkit is still work in progress, we implement several algorithms that have never been implemented before, such as
SSL-ALM \cite{JakubReview},
Stochastic Switching Subgradient \cite{HuaLin2023} 
and the Stochastic Ghost \cite{FacKun2023}.

\textbf{Our contributions.}
The contributions of this paper are:
\begin{itemize}[nolistsep]
    \item a literature review of algorithms subject to handling \eqref{eq:pb};
    \item a toolbox that implements algorithms applicable in constrained training of DNNs;
    \item numerical experiments that compare these algorithms on a real-world dataset.
\end{itemize}

\section{Algorithms}
\label{sec:algorithms}

Solving \eqref{eq:pb} encounters the following challenges:
\begin{itemize}[nolistsep]
    \item large-scale objective and constraint functions, which require sampling schemes,
    \item the necessity of incorporating inequality constraints, not merely equality constraints,
    \item the necessity to cope with the nonconvexity and nonsmoothness, due to the nature of neural networks.
\end{itemize}
In this section, we identify the algorithms that address these challenges 
most precisely. However, we note that there exists currently no algorithm with guarantees for such a~general setting.

\paragraph{Notation.}

We denote the projection of a~point $x$ onto a~set $\cX$ by
$\proj_\cX(x) = \arg\min_{v\in\cX}\|x-v\|^2$.
We distinguish between the random variable $\xi$ associated with the objective function
and the random variable $\zeta$ associated with the constraint function.
We denote $\cP_\xi$ and $\cP_\zeta$ their probability distributions.

\subsection{Review of methods for constrained ERM}
We compare recent constrained optimization algorithms 
considering a~stochastic objective function in Table~\ref{table:algs_asm}.
We note that most of them
do not consider the case of stochastic constraints.
Among those which do consider stochastic constraints,
only three admit inequality constraints.
Moreover, with the exception of \cite{HuaLin2023}, all the algorithms
in Table~\ref{table:algs_asm} assume $F$ to be at least $\C^1$, which
makes addressing the challenge of nonsmoothness of $F$ infeasible.
The recent paper \cite{DacDruKakLee2018} leads us to the conclusion
that assuming the objective and constraint functions to be tame
and locally Lipschitz is a~suitable requirement for solving \eqref{eq:pb}
with theoretical guarantees of convergence. 
At this point, however, no such algorithm exists, 
to the best of our knowledge.
Consequently, we consider the practical performance of the algorithms that address 
the challenges of solving \eqref{eq:pb} most closely: 
SSL-ALM \cite{JakubReview}, and Stoch. Switching Subgradient~\cite{HuaLin2023}.

\subsection{Stochastic Smoothed and Linearized AL Method (SSL-ALM)}\label{sec:SSL-ALM}
The Stochastic Smoothed and Linearized AL Method (SSL-ALM)
was described in \cite{JakubReview} for optimization
problems with stochastic linear constraints.
Although problem \eqref{eq:pb} has
non-linear inequality constraints, we use the SSL-ALM due to the lack
of algorithms in the literature dealing with stochastic non-linear constraints;
see Table~\ref{table:algs_asm}.
The transition between equality and inequality constraints can be handled by either taking the maximum between the constraint value and 0 or using slack variables.
Following the structure of \cite{JakubReview}, we minimize over the set $\cX=\R^n\times\R_{\geq 0}^m$.
The method is based on the augmented Lagrangian (AL) function 
$L_\rho(x,y)=F(x) + y^\top C(x)+ \frac{\rho}{2}\|C(x)\|^2$,
which is a~result of merging the Lagrange function with 
the penalty methods \cite{Bertsekas2014}. 
Adding a smoothing term and a variable $z \in \bbR^n$ yields the proximal AL function
\[
    K_{\rho,\mu}(x,y,z) = L_\rho(x,y) + \frac{\mu}{2}\|x-z\|^2.
\]
The SSL-ALM method was originally proposed in \cite{JakubReview}
where it is interpreted
as an inexact gradient descent step on the Moreau envelope. An important property of the Moreau envelope is that its stationary
points coincide with those of the original function.

In each iteration, we sample $\xi\iid\mathcal{P}_\xi$ to evaluate the objective
and $\zeta_1$, $\zeta_2\iid\mathcal{P}_\zeta$ to evaluate the constraint function
and its Jacobian matrix, respectively. The function
\begin{equation}\label{eq:defG}
    G(x, y, z; \xi, \zeta_1, \zeta_2) = \nabla f(x, \xi) + \nabla c (x, \zeta_1)^\top y + \rho \nabla c (x, \zeta_1)^\top c(x, \zeta_2) + \mu (x - z)
\end{equation}
is defined so that, in iteration $k$, 
$\bbE_{\xi,\zeta_1,\zeta_2}[G(x_k, y_{k+1}, z_k; \xi, \zeta_1, \zeta_2)]
= \nabla K_{\rho,\mu}(x_k,y_{k+1},z_k)$.
Omitting some details, the updates are performed using some
parameters $\eta, \tau$, and $\beta$ as follows:
\begin{equation}\label{eq:sAL}
    \begin{aligned}
      &y_{k+1} = y_{k}+\eta c(x, \zeta_1), \\
      &x_{k+1} = \proj_{\cX}(x_k - \tau G(x_k, y_{k+1}, z_k; \xi, \zeta_1, \zeta_2)), \\
      & z_{k+1} = z_k + \beta (x_k-z_k).
    \end{aligned}
\end{equation}

\subsection{Stochastic Switching Subgradient Method (\SSW)}\label{sec:SSW}
The Stochastic Switching Subgradient method was described in \cite{HuaLin2023} for optimization
problems over a~closed convex set $\cX\subset\bbR^d$ which is easy to project on and for
weakly convex objective and constraint functions $F$ and $C$ which may be non-smooth.
This is why the notion of gradient of $F$ and $C$ is replaced by a~more general notion
of subgradient, which is an element of a~subdifferential.

The algorithm requires as input a~prescribed sequence of infeasibility
tolerances $\epsilon_k$ and sequences of stepsizes $\eta^f_k$ and $\eta^c_k$. 
In iteration $k$, we sample
$\zeta_1,\ldots,\zeta_J\iid\mathcal{P}_\zeta$
to compute an estimate $\overline{c}\strut^{J}(x_k)$. 
If $\overline{c}\strut^{J}(x_k)$ is smaller than $\epsilon_k$,
we sample $\xi\iid\mathcal{P}_\xi$ and
an update between $x_k$ and $x_{k+1}$
is computed using a~stochastic estimate $S^f(x_k,\xi)$ of an element of the subdifferential
$\partial F(x_k)$
of the objective    function:
\[
    x_{k+1} = \proj_{\cX} (x_k-\eta^f_{k} S^f(x_k,\xi) ).
\]
Otherwise, we sample $\zeta\iid\mathcal{P}_\zeta$ 
and the update is computed using 
a~stochastic estimate $S^c(x_k,\zeta)$ of an element of the subdifferential $\partial C(x_k)$
of the constraint function:
\[
    x_{k+1} = \proj_{\cX} (x_k-\eta^c_{k} S^c(x_k,\zeta) ).
\]

The algorithm presented here is slightly more general than the one
presented in \cite{HuaLin2023}: we allow for the possibility of different stepsizes for the objective update, $\eta_k^f$, and the constraint update $\eta_k^c$, while the original method employs equal stepsizes $\eta_k^f = \eta_k^c$.

\subsection{Package implementation}

In the package, both algorithms are implemented with an API close to that of a PyTorch Optimizer. In addition to the \texttt{step()} method, which performs the primal update, we add a \texttt{dual\_step()} method, which updates the dual parameters and performs other related tasks. This allows for pipelines similar to those of PyTorch.

\begin{table}[t]
  \centering
  \caption{
    Assumptions on objective and constraint functions, $F$ and $C$, 
    which allow for theoretical convergence proofs.
    \label{table:algs_asm}
  }
  \footnotesize
  \resizebox{0.8\textwidth}{!}{
  \begin{tabular}{lcclc|ccc|cclc}
    \toprule
      & \multicolumn{4}{c}{Objective function $F$} & \multicolumn{7}{c}{Constraint function $C$} \\ \cmidrule(lr){2-5} \cmidrule(lr){6-12}
      Algorithm& \rot{90}{stochastic}  
      & \rot{90}{weakly convex} & \rot{90}{$\C^1$ with Lipschitz $\nabla F$}
      & \rot{90}{tame loc. Lipschitz} & \rot{90}{stochastic}  & 
      \rot{90}{$C(x)=0$} & \rot{90}{$C(x) = 0$ and $C(x)\le 0$} & \rot{90}{linear} & \rot{90}{weakly convex} &  \rot{90}{$\C^1$ with Lipschitz $\nabla C$}
      & \rot{90}{tame loc. Lipschitz}  \\ \midrule
      SGD &\cmark&(\cmark)&(\cmark)&\cmark& &&&&&& \\ \midrule
    \cite{BerCur2023} \cite{FaNaMaKo2024} \cite{Curtis2024}&  \cmark&--&\cmark&--&--&\cmark& -- &-- &--&\cmark&-- \\
    \cite{NaAnKo2022}   &  \cmark&--&\cmark ($\C^3)$&--&--&\cmark& --&--&--&\cmark ($\C^3)$&--\\
    \cite{ShiWaWa2022} \cite{CurRob2024}& \cmark&--&\cmark&--&--&(\cmark)&\cmark&--&--&\cmark&-- \\
    \cite{NaAnKo2023}   &  \cmark&--&\cmark ($\C^2$)&--&--&(\cmark)&\cmark&--&--&\cmark ($\C^2$)&--\\
    \cite{Bolla2023}    &  \cmark & -- &\cmark (+ cvx)&--&--&\cmark& --&\cmark&--&--&-- \\
    \cite{Ozto2023}     &  \cmark&--&\cmark&--&\cmark& \cmark & --&--&--&\cmark&-- \\
    SSL-ALM \cite{JakubReview} &\cmark&--&\cmark&--& \cmark &(\cmark)&\cmark&\cmark&--&--&-- \\
    Stoch. Ghost \cite{FacKun2023} & \cmark&--&\cmark&--&\cmark& (\cmark) &\cmark&--&--&\cmark&-- \\
    Stoch. Switch. Subg. \cite{HuaLin2023} & \cmark & \cmark &--&--&\cmark& (\cmark) &\cmark&--&\cmark&--&--\\
    \bottomrule
  \end{tabular}
  }
  \vspace{-2ex}
\end{table}

\section{Experimental evaluation}
\label{sec:experiments}

\paragraph{Data.}
In this section, we illustrate the presented algorithms on a real-world instance of the ACS dataset, which is  based on the ACS PUMS data sample (American Community Survey
Public Use Microdata Sample), accessed in Python via the Folktables package \cite{ding2021retiring}.
In particular, we use the ACSIncome dataset over the state of Virginia, and choose the
binary classification task of predicting whether an individual's income is over \$50,000.
The dataset contains 9 features
and 46,144 data points.
We choose marital status (\textbf{MAR}) as the protected attribute, which is a categorical attribute with 5 values: \texttt{Married, Widowed, Divorced, Separated}, and \texttt{Never Married/Under 15}. The dataset is split randomly into train (80\%, 36,915 points) and test (20\%, 9,229 points) subsets. The protected attribute is then removed from the data so that the model cannot learn from it directly.
The data is normalized using Scikit-Learn \texttt{StandardScaler}.

Using a utility provided with the humancompatible.train package, we sample equal number of data points from each subgroup at each iteration to ensure that all constraints can be computed.

\paragraph{Problems.}
In all problems, we set the objective function the Binary Cross Entropy with Logits Loss
\begin{equation}\label{eq:BCE}
    \ell( f_\theta(X_i) , Y_i) = -Y_i\cdot \log \sigma(f_\theta(X_i))-(1-Y_i)\cdot\log(1-\sigma(f_\theta(X_i))),
\end{equation}
where $\sigma(z) = \frac{1}{1+e^{-z}}$ is the sigmoid function, and
the prediction function $f_\theta$ is a neural network with 2 interconnected hidden layers of sizes 64 and 32 and ReLU activation, with a total of 194 parameters.

For the constrained algorithms, for each of the 5 groups (defined by the value of the protected attribute), we put an upper bound on the absolute difference between Positive Rate computed on that group and the overall Positive Rate:
\begin{equation}\label{eq:PRConstraint}
    \lvert P(f_\theta(X) = 1 \mid G = g) - P(f_\theta(X) = 1)\rvert\leq c,
\end{equation}
where $G$ is the group membership indicator.

We set the constraint bound $c = 0.05$ for all groups; we use Fairret \cite{buyl2024fairret} to compute the constraints.

\paragraph{Algorithms and parameters.}
We compare the performance of two algorithms for solving the constrained problem:  \emph{(1)} SSL-ALM (Sec. \ref{sec:SSL-ALM} - parameters $\mu = 2.0$, $\rho = 1.0$, $\tau = 0.05$, $\eta = 0.1$, $\beta = 0.5$, $M_y=100$), %
and \emph{(2)} Stochastic Switching Subgradient (\SSW{}) (Sec. \ref{sec:SSW} - $\eta^f_k = 0.01$, $\eta^c_k = 0.01$, 
$\epsilon_1 = 10^{-1}$, $\epsilon_k = \frac{\epsilon_{k-1}}{\sqrt{k}}$ for all $k \geq 1$). As a baseline, we also train the network without constraints using Adam with default hyperparameter values found in PyTorch.

\paragraph{Setup.}
Experiments are conducted on an Asus Zenbook UX535 laptop with AMD Ryzen 7 5800H CPU, and 16GB RAM. Each algorithm is run for 1 minute, repeated 5 times.

\begin{figure}[t]
  \begin{subfigure}{\textwidth}
    \includegraphics[width=\linewidth]{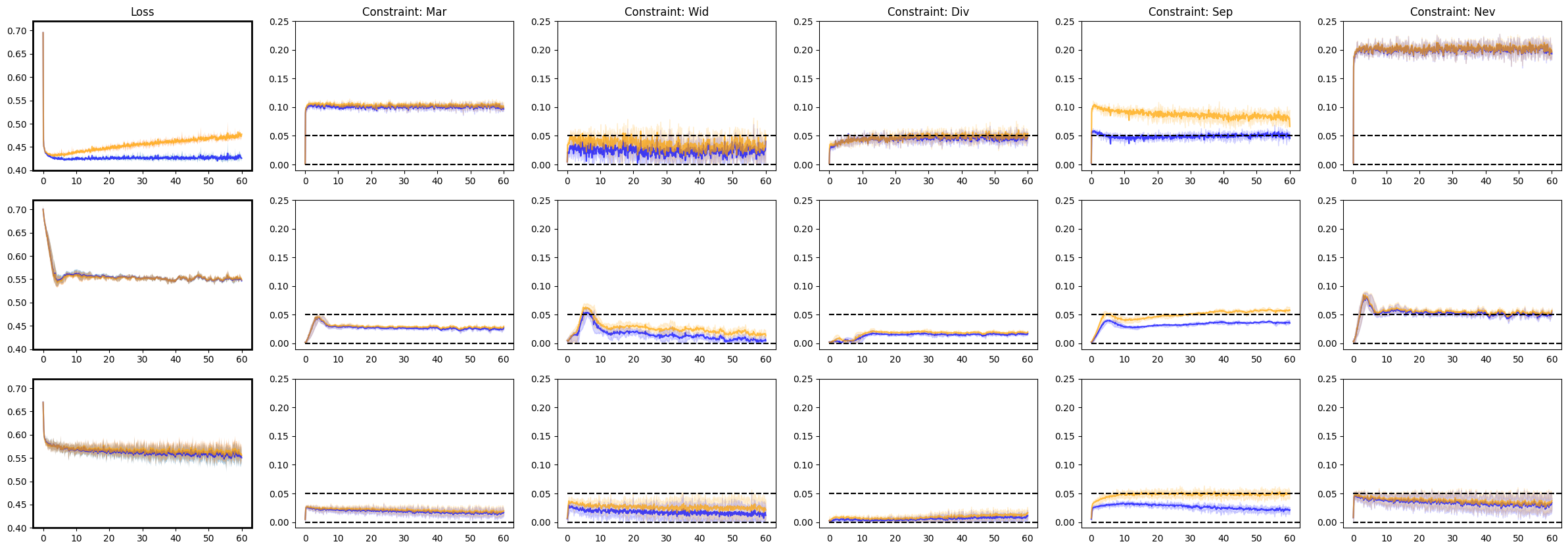}
  \end{subfigure}
  \caption{Train (blue) and test (orange) statistics over time (s) on the ACS \textcolor{black}{Income} dataset for each algorithm: Adam (top-row), SSL-ALM (middle-row), and SSw (bottom-row). The plots depict the mean values for loss (leftmost column) and constraints (second to rightmost column) at each timestamp (rounded to the nearest 0.1 seconds) over 5 runs. The shaded area depicts the region between the lowest and the highest value for corresponding statistics calculated the same way. \label{fig:opt_train}}
  \vspace{-2ex}
\end{figure}

\paragraph{Optimization performance.}
\Cref{fig:opt_train} presents the evolution of loss and constraint values over the train and test datasets for the baseline (Adam) (row 1), and the two algorithms addressing the constrained problem (SSL-ALM on row 2 and SSw on row 3). 
Both SSL-ALM and SSw algorithms succeed in keeping the constraint values within bounds, with SSL-ALM minimizing the objective function faster under the chosen hyperparameters.

\subsubsection*{Acknowledgements}
The work was funded by European Union’s Horizon Europe research and innovation program under grant agreement No. 101070568.

\bibliographystyle{plain} %
\bibliography{references.bib}

\end{document}